\pdfoutput=1

\documentclass[11pt]{article}

\usepackage[]{EMNLP2023}

\usepackage{times}
\usepackage{latexsym}

\usepackage[T1]{fontenc}

\usepackage[utf8]{inputenc}

\usepackage{microtype}

\usepackage{inconsolata}

\usepackage{graphicx}
\usepackage{amssymb}
\usepackage{amsmath}
\usepackage{amsthm}
\usepackage{booktabs}
\usepackage{algorithm}
\usepackage{algorithmic}
\usepackage{placeins}
\usepackage{multirow}
\usepackage{ulem}
\usepackage{subfigure}
\usepackage{enumitem}
\newcommand{\hide}[1]{}

\newtheorem{problem}{Problem}

\title{Precedent-Enhanced Legal Judgment Prediction with LLM and Domain-Model Collaboration}

\author{Yiquan Wu$^1$, Siying Zhou$^1$, Yifei Liu$^1$, Weiming Lu$^{1*}$, Xiaozhong Liu$^{2}$\\
\textbf{Yating Zhang$^{3}$, Changlong Sun$^{13}$, Fei Wu$^{1*}$, Kun Kuang$^{1*}$}\\
$^1$Zhejiang University, Hangzhou, China\\
$^2$Worcester Polytechnic Institute, USA\\
$^3$Alibaba Group, Hangzhou, China\\
\small\texttt{\{wuyiquan, zhousiying, liuyifei, luwm, kunkuang\}@zju.edu.cn, yatingz89@gmail.com}\\
\small\texttt{xliu14@wpi.edu, changlong.scl@taobao.com, wufei@cs.zju.edu.cn}}

\begin{document}
\maketitle
\begin{abstract}

Legal Judgment Prediction (LJP) has become an increasingly crucial task in Legal AI, i.e., predicting the judgment of the case in terms of case fact description. 
Precedents are the previous legal cases with similar facts,
which are the basis for the judgment of the subsequent case in national legal systems.
Thus, it is worthwhile to explore the utilization of precedents in the LJP.
Recent advances in deep learning have enabled a variety of techniques to be used to solve the LJP task. These can be broken down into two categories: large language models (LLMs) and domain-specific models. LLMs are capable of interpreting and generating complex natural language, 
while domain models are efficient in learning task-specific information. 
In this paper, we propose the precedent-enhanced LJP framework (PLJP) – a system that leverages the strength of both LLM and domain models in the context of precedents. 
Specifically, the domain models are designed to provide candidate labels and find the proper precedents efficiently, and the large models will make the final prediction with an in-context precedents comprehension. Experiments on the real-world dataset demonstrate the effectiveness of our PLJP. Moreover, our work shows a promising direction for LLM and domain-model collaboration that can be generalized to other vertical domains.

\end{abstract}

\renewcommand{\thefootnote}{\fnsymbol{footnote}}
\footnotetext[1]{Corresponding Authors.}
\renewcommand{\thefootnote}{\arabic{footnote}}

\section{Introduction}

Legal AI has been the subject of research for several decades, with the aim of assisting individuals in various legal tasks, including legal QA \citep{monroy2009nlp}, court view generation \citep{wu2020biased}, legal entity recognition \citep{cardellino2017legal}, and so on. As one of the most important legal tasks, legal judgment prediction (LJP) aims to predict the legal judgment of the case based on the case fact description. The legal judgment typically includes the law article, charge and prison term.

\begin{figure}[t]
\centering
\includegraphics[width=0.45\textwidth]{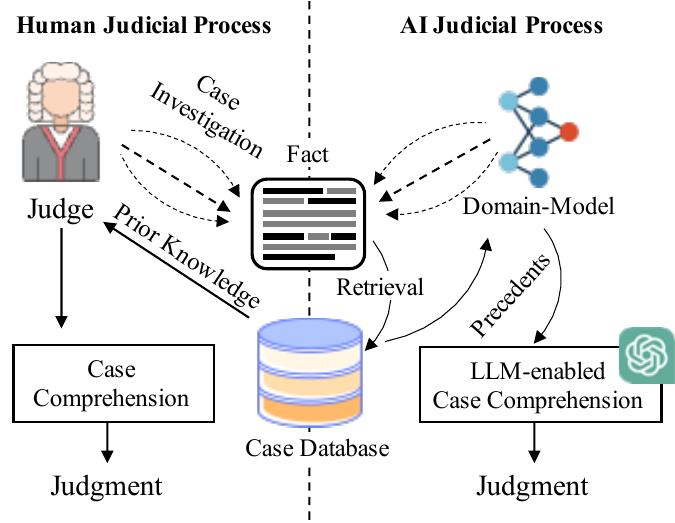}
\caption{
An illustration of the judicial process, our motivation is to promote the collaboration between the domain model and LLM (right part) for simulating the judicial process of the human judge (left).
}
\label{fig:example}
\vspace{-10pt}
\end{figure}

Precedents, which refer to previous cases with similar fact descriptions, hold a crucial position within national legal systems \citep{guillaume2011use}. On a more macro level, precedents are known as the collective body of judge-made laws in a nation\citep{Bryandict}. They serve the purpose of ensuring consistency in judicial decisions, providing greater legal guidance to judges and facilitating legal progress and evolution to meet dynamic legal demands. In the Common Law system, the precedents are the mandatory basis of the judgment of the subsequent case \citep{rigoni2014common}. In the Civil Law system, judge-made laws are perceived as secondary legal sources while written laws are the basic legal sources\citep{LarenzMethod}. In the contemporary era, there is also a growing trend to treat the precedents as a source of ``soft'' law \citep{fon2006judicial}, and judges are expected to take them into account when reaching a decision \citep{guillaume2011use}. Thus, it is worthwhile to explore the utilization of precedents in the legal judgment prediction.

With the development of deep learning, many technologies have been adopted in the LJP task, which can be split into two categories: large language models (LLMs) and domain-specific models \citep{ge2023openagi}.
Owing to extensive training, LLMs are good at understanding and generating complex natural language, as well as in-context learning. On the other hand, domain-specific models are designed to cater to specific tasks and offer cost-effective solutions.
However, when it comes to incorporating precedents into the LJP task, both categories of models face certain limitations. LLMs, constrained by their prompt length, struggle to grasp the meaning of numerous abstract labels and accurately select the appropriate one. For domain models, though trained with label annotations, the drawback is the limited ability to comprehend and distinguish the similarities and differences between the precedents and the given case.

In this paper, as Fig. \ref{fig:example} shows, we try to collaborate the LLMs with the domain-specific models and propose a novel precedent-enhanced legal judgment prediction framework (PLJP). Specifically, domain models contribute by providing candidate labels and finding the proper precedents from the case database effectively; the LLMs will decide the final prediction through an in-context precedent comprehension. 

Following the previous LJP works \citep{zhong2018legal,yue2021neurjudge, dong2021legal}, our experiments are conducted on the publicly available real-world legal dataset. 
To prevent any potential data leakage during the training of the LLMs, where the model may have already encountered the test cases, we create a new test set comprising cases that occurred after 2022. This is necessary because the LLMs we utilize have been trained on a corpus collected only until September 2021. By doing so, we ensure a fair evaluation of the PLJP framework. Remarkably, our proposed PLJP framework achieves state-of-the-art (SOTA) performance on both the original test set and the additional test set.

To sum up, our main contributions are as follows:
\begin{itemize}
    \item We address the important task of legal judgment prediction (LJP) by taking precedents into consideration.

    \item We propose a novel precedent-enhanced legal judgment prediction (PLJP) framework that leverages the strength of both LLM and domain models.

    \item We conduct extensive experiments on the real-world dataset and create an additional test set to ensure the absence of data leakage during LLM training. The results obtained on both the original and additional test sets validate the effectiveness of the PLJP framework. 

    \item Our work shows a promising direction for LLM and domain-model collaboration that can be generalized over vertical domains. We make all the codes and data publicly available to motivate other scholars to investigate this novel and interesting research direction\footnote{The github link is hidden for the anonymous review.}.
\end{itemize}

\section{Related Work}
\subsection{Legal AI}
Legal Artificial Intelligence (Legal AI) aims to enhance tasks within the legal domain through the utilization of artificial intelligence techniques \citep{zhong2020does, katz2023natural}. Collaborative efforts between researchers in both law and computer fields have been lasting to explore the potential of Legal AI and its applications across various legal tasks. These tasks encompass areas such as legal question answering (QA) \citep{monroy2009nlp}, legal entity recognition \citep{cardellino2017legal}, court view generation \citep{wu2020biased}, legal summarization \citep{hachey2006extractive, bhattacharya2019comparative}, legal language understanding\citep{chalkidis2021lexglue} and so on.

In this work, we focus on the task of legal judgment prediction, which is one of the most common tasks in Legal AI.

\subsection{Legal Judgment Prediction}
Legal judgment prediction (LJP) aims to predict judgment results based on the fact descriptions automatically \citep{lin2012exploiting,chalkidis2019neural, yue2021neurjudge, xu2020distinguish,niklaus2021swiss,malik2021ildc, feng2022legal, lyu2022improving, zhang2023contrastive}. 
The LJP methods in earlier years required manually extracted features \citep{keown1980mathematical}, which is simple but costly. 
Owing to the prosperity of machine learning \citep{WuWZXCZ022, shen2022mask, li2022end, li2022hero, ijcai2022p654, li2023winner, zhang2023data}, researchers began to formalize the LJP problem with machine learning methods. These data-driven methods can learn the features with far less labor (e.g., only the final labels are required).
\citet{sulea2017exploring} developed an ensemble system that averages the output of multiple SVM to improve the performance of LJP. \citet{luo2017learning} utilized an attention mechanism in the LJP. 
\citet{zhong2018legal} considered the dependency of the sub-tasks in the LJP. 
\citet{yue2021neurjudge} investigated the problem by separating the representation of fact description into different embedding. \citet{liu2022augmenting} used contrastive learning in the LJP.

However, these existing LJP methods tend to overlook the significance of precedents. In this study, we propose a precedent-enhanced LJP framework (PLJP) that leverages the collaboration between domain-specific models and large language models (LLMs) to address the LJP task.

\subsection{Precedent Retrieval}
The precedent is the basis of judgment in the Common Law system, and also an important reference for decision-making in the Civil Law system. Therefore, precedent retrieval is another valuable task in Legal AI \citep{althammer2021dossier}. 
There are two main precedent retrieval models: expert knowledge-based models and natural language processing (NLP)-based models \citep{bench2012history}. Expert knowledge-based models use the designed sub-elements to represent the legal cases \citep{saravanan2009improving}, while NLP-based models mainly convert the text into embeddings and then calculate the similarity from the embedding level \citep{ma2021retrieving, chalkidis2020legal}.

Most retrieval models required additional annotation so can not be directly applied to the LJP task. In our paper, we use an unsupervised dense retrieval model \citep{izacard2022unsupervised} to get the precedents, which can be updated by other retrieval models if needed.

\subsection{Large Language Models}
Large language models (LLMs), such as ChatGPT, have attracted widespread attention from society \citep{zhao2023survey}. With pre-training over large-scale corpora, LLMs show strong capabilities in interpreting and generating complex natural language, as well as reasoning (e.g., in-context learning). The technical evolution of LLMs has been making an important impact on the fields of natural language processing \citep{brown2020language, touvron2023llama}, computer vision \citep{shao2023prompting, wu2023visual}, and reinforcement learning \citep{du2023guiding}. In the legal domain, LLMs can also be used for many tasks such as legal document analysis and legal document writing \citep{sun2023short}.

However, in the prediction tasks, which can involve dozens of abstract labels, the performance of LLMs is not as good as in generation tasks, due to the limited prompt length. In this paper, we explore the utilization of LLMs in the LJP task with the collaboration of domain-specific models.

\section{Problem Formulation}
In this work, we focus on the problem of legal judgment prediction. We first clarify the definition of the terms as follows.

$\bullet$ \textbf{Fact Description} refers to a concise narrative of the case, which typically includes the timeline of events, the actions or conduct of each party, and any other essential details that are relevant to the case. Here we define it as a token sequence $f=\{w_t^f\}_{t=1}^{l_f}$, where $l_f$ is the length.

$\bullet$ \textbf{Judgment} is the final decision made by a judge in a legal case based on the facts and the precedents. It typically consists of the law article, the charge, and the prison term. We represent the judgment of a case as $j = (a, c, t)$, where $a$, $c$, $t$ refer to the labels of article, charge and prison term, respectively.

$\bullet$ \textbf{Precedent} is the previous case with a similar fact. The judgments of the precedents are important references for the current case. Here, a precedent is defined as $p = (f_p, j_p)$, where $f_p$ is its fact description and $j_p$ is its judgment. For a given case, there can be several precedents, which can be denoted as $P=\{p_1, p_2, ...,p_n\}$, where $n$ is the number of precedents.

Then the problem can be defined as:
\begin{problem} [Legal Judgment Prediction]
Given the fact description $f$, our task is to get and comprehend the precedents $P$, then predict the judgment $j = (a, c, t)$.
\end{problem}

\section{Precedent-Enhanced LJP (PLJP)}
\begin{figure*}[ht]
    \centering
    \includegraphics[scale = 0.55]{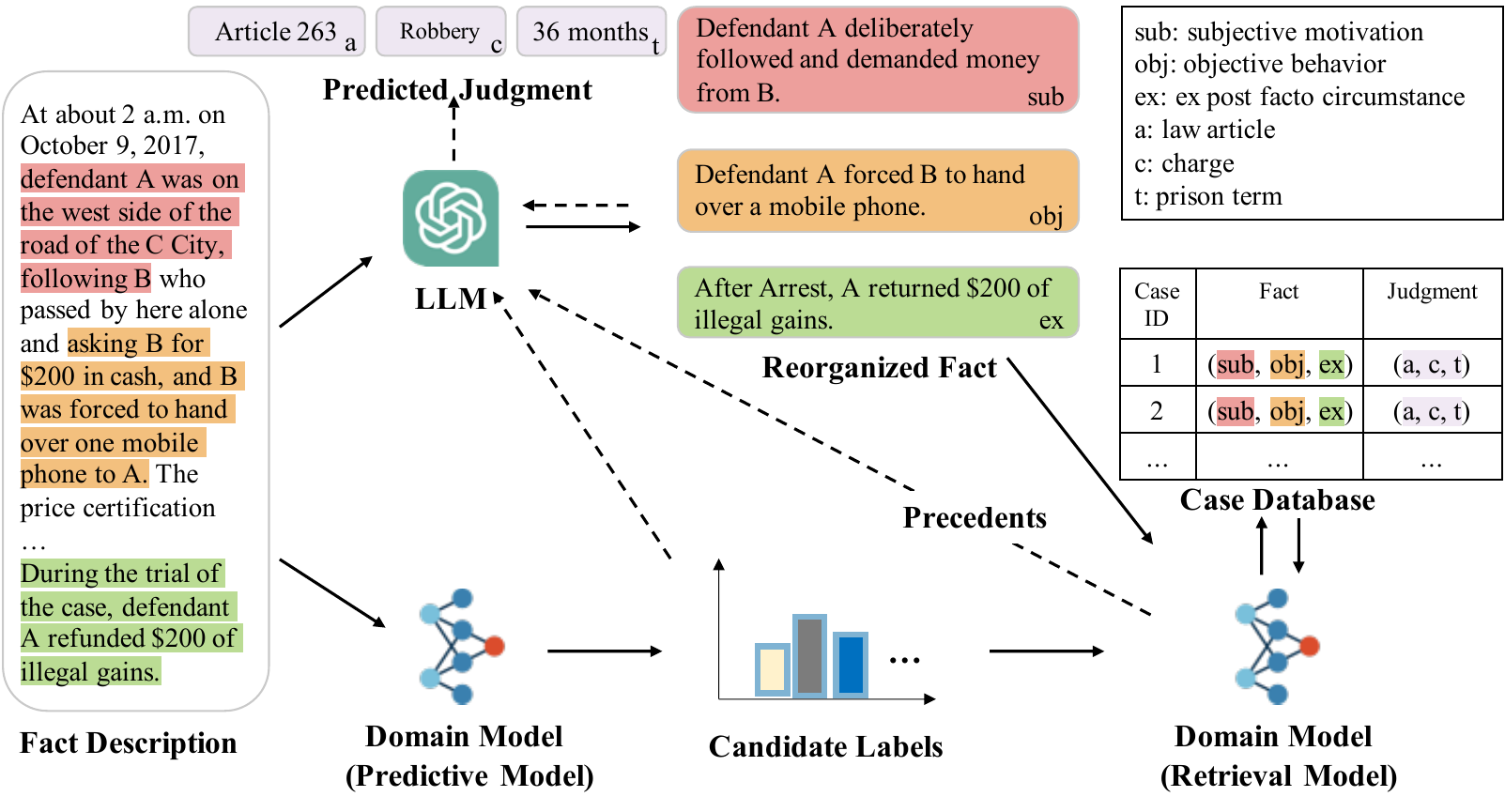}
    \caption{The overall framework of PLJP, where the sub, obj and ex refer to the subjective motivation, objective behavior and ex post facto circumstance, respectively. The solid lines are the precedent retrieval process, while the dotted lines represent the process of the prediction.}
    \vspace{-10pt}
    \label{fig:modelarch}
\end{figure*}

In this section, we describe our precedent-enhanced legal judgment prediction framework (PLJP), Fig. \ref{fig:modelarch} shows the overall framework.

\subsection{Case Database Construction}
Before we use the precedents, we have to collect a large number of previous cases to construct a case database.
Since the fact descriptions are usually long and elaborate, it is difficult for the models to get the proper precedents. To this end, we reorganize the fact description of these previous cases with the help of LLMs.
\subsubsection{Fact Reorganization} \label{sec:fo}
Given a fact description of a case, we summarize it from three aspects: subjective motivation, objective behavior, and ex post facto circumstances. The reorganization doesn't require human annotation and is completed by the LLMs with the following prompts: \textit{
``A fact description can be categorized into subjective motivation, objective behavior, and ex post facto circumstances. 
Subjective motivation refers to the psychological attitude of the perpetrator towards their harmful actions and their consequences, including intent, negligence, and purposes of the crime. 
Objective behavior pertains to the necessary conditions for constituting a crime in terms of observable activities, including harmful conduct, harmful results, and the causal relationship between the conduct and the results. 
Ex post facto circumstances are various factual situations considered when determining the severity of penalties. Mitigating circumstances for lenient punishment include voluntary surrender and meritorious conduct, while aggravating circumstances for harsher punishment include recidivism.
Based on the provided information, your task is to summarize the following facts.''
}

The reorganization reduces the length of facts and makes the precedents easy to get and comprehend in the PLJP.

After the reorganization, the fact description $f$ is translated to a triplet $(sub, obj, ex)$, which indicates the subjective motivation, objective behavior, and ex post facto circumstances, respectively. 
Finally, a previous case in the case database is stored as a pair of reorganized facts and the judgment.

\subsection{Legal Judgment Prediction}
Next, we describe the collaboration of the LLM and domain models in legal judgment prediction.

\subsubsection{Domain Models}
The domain models are trained on specific datasets, aiming to solve certain tasks. Here, we use two kinds of domain models, including the predictive model and the retrieval model.
\paragraph{Predictive model.} The predictive model takes the fact description as the input and outputs the candidate labels of the three sub-tasks (e.g., law article, charge, prison term).
Since the fact description $f=\{w_t^f\}_{t=1}^{l_f}$ are sequences of words, we first transform it into embedding sequence  $H^f \in \mathbb{R}^{l_f \times d}$ with an Encoder:
\begin{equation}
    H^f = \operatorname{Encode}(f),
\end{equation}
where $H^f = {h^f_1, h^f_2, ..., h^f_{l_f}}$, and $d$ is the dimension of the embedding.

We take a max-pooling operation to obtain the pooled hidden vector $h^f \in \mathbb{R}^{d}$ and then feed it into a fully-connected network with softmax activation to obtain the label probability distribution $P \in \mathbb{R}^{m}$:
\begin{equation}
\begin{aligned}
    h^f &= \operatorname{MaxPooling}(H^f),\\
    P &= \operatorname{Softmax}(W^p \cdot h^f + b^p),
\end{aligned}
\end{equation}
where $W^p \in \mathbb{R}^{m \times d}$ and $b^p \in \mathbb{R}^{m}$ are learnable parameters. Note $m$ varies in different sub-tasks.

Then, each sub-task gets its candidate labels according to the probability distribution $P$, and the number of candidate labels is equal to the number of precedents $n$.

\paragraph{Retrieval model.} The retrieval model aims to get the proper precedents of the given case based on its reorganized fact $(sub, obj, ex)$.

Formally, to get the similarity score of any two texts $D_1$ and $D_2$, we will first encode each of them independently using the same encoder:
\begin{equation}
    h_{D_1} = \operatorname{Encoder}(D_1),
    h_{D_2} = \operatorname{Encoder}(D_2),
\end{equation}
where $h_{D_1} \in \mathbb{R}^{d'}$ and $h_{D_2} \in \mathbb{R}^{d'}$ are the embedding of each, $d'$ is the dimension. 
The similarity score $s(D_1, D_2)$ is then the cosine  similarity of the $h_{D_1}$ and $h_{D_2}$:
\begin{equation}
    s(D_1, D_2) =  \frac{h_{D_1} \cdot h_{D_2}}{\left\| h_{D_1} \right\| \left\| h_{D_2} \right\|}.
\end{equation}

Here we concatenate the $sub$, $obj$ and $ex$ into a whole text to calculate the similarity score of the given case and the cases in the case database.

For each candidate label, we pick one case as the precedent: the case that has the highest similarity score and has the same label. For example, if the label ``Theft'' is in the candidate labels in the charge prediction, we will find the most similar previous case with the same label as the corresponding precedent. The one-to-one relationship between the candidate label and precedent helps the LLM distinguish the differences among the labels. In other words, the precedent serves as a supplementary explanation of the label.

Finally, we get precedents $P = \{p_1, p_2, ..., p_n\}$ for the given case.

\subsubsection{LLMs}
The large language models are models with billions of parameters, which are trained on large-scale corpora, and show strong capabilities in interpreting and generating complex natural language. LLMs contribute to PLJP by fact reorganization and in-context precedent comprehension.
\paragraph{Fact Reorganization}
The fact reorganization is described in case database construction (Sec. \ref{sec:fo}), which aims to summarize the fact description from three aspects by the LLMs. Besides the database contribution, as Fig. \ref{fig:modelarch} shows, when a new test case comes, the LLMs will reorganize the fact description with the same prompt.

\paragraph{In-Context Precedent Comprehension}
Since LLMs are capable of understanding complex natural language, we stack the given case with its precedents and let the LLMs make the final prediction by an in-context precedent comprehension.
Specifically, the prompt of law article prediction is designed as follows:
\textit{
``Based on the facts, we select the candidate law articles by the domain models and select the following three precedents based on the candidate law articles. Please comprehend the difference among the precedents, then compare them with the facts of this case, and choose the final label.''
}

Consider the topological dependencies among the three sub-tasks \citep{zhong2018legal}, in the prediction of charge, we add the predicted law article in the prompt; in the prediction of prison term, we add the predicted law article and charge.

\subsection{Training}
In PLJP, considering the realizability, we train domain models on legal datasets and leave the LLMs unchanged. To train predictive models, the cross-entropy loss is employed. As for retrieval models, contrastive loss is used like \citet{izacard2022unsupervised}.

\section{Experiments}
\begin{table}[ht]
\centering
\small
\begin{tabular}{lcc}
\hline
\textbf{Type}                      & \textbf{CAIL2018} & \textbf{CJO22} \\ \hline
\# Law Article                     & 164                &   164             \\
\# Charge                          & 42                &   42          \\
\# Prison Term                     &  10              &   10         \\
\# Sample                          &  82138             &  1698        \\
Avg. \# words in Fact             &  288.6           &  461.7        \\ \hline
\end{tabular}
\caption{Statistics of datasets.}
\label{tab:stat}
\end{table}

\begin{table*}[ht]
\small
    \centering
\begin{tabular}{lcccc|cccc}
\hline
\multicolumn{1}{c}{\multirow{2}{*}{Method}} & \multicolumn{4}{c|}{CJO22} & \multicolumn{4}{c}{CAIL2018} \\
\multicolumn{1}{c}{}                        & Acc   & Ma-P  & Ma-R  & Ma-F  & Acc  & Ma-P & Ma-R & Ma-F \\ \hline
CNN \citep{lecun1989backpropagation}        & 76.14 & 35.48 & 38.55 & 35.39 & 80.50 & 40.10 & 38.33 & 38.49\\
BERT \citep{devlin2018bert}                 & 82.62 & 45.89 & 47.91 & 45.83 & 82.77 & 36.82 & 35.94 & 35.82  \\
Roberta \citep{liu2019roberta}              & 80.32 & 42.36 & 44.22 & 41.80 & 83.08 & 48.09 & 44.25 & 44.87  \\
TopJudge \citep{zhong2018legal}             & 78.73 & 40.38 & 41.47 & 40.09 & 80.46 & 40.96 & 40.96 & 38.24 \\
R-Former \citep{dong2021legal}              & \underline{87.69} & 53.03 & 49.35 & 50.23 &  \textbf{87.82}  & 56.13 & \underline{56.57} & \underline{55.81}  \\
LADAN \citep{xu2020distinguish}             & 79.44 & 48.43 & 44.13 & 46.18 & 82.82 & 42.57 & 39.00 & 40.71  \\
NeurJudge \citep{yue2021neurjudge}          &  71.38   &  52.86 & 53.52 & 52.62   & 76.91 &  55.95 & 52.92 & 53.56 \\ 
EPM\citep{feng2022legal} & 84.19 & 47.21 & 43.79 & 44.39 & 85.80 & 49.08 & 45.76 & 47.32 \\ 
CTM\citep{liu2022augmenting} & 79.44 & 47.83 & 42.25 & 43.43 & 84.72 & 46.46 & 44.83 & 45.10 \\ 
Dav003          & 2.10 & 0.82 & 0.17 & 0.26 & 1.02 & 0.30 &  0.08 & 0.13 \\ 
3.5turbo         & 9.13 & 2.54 & 1.61 & 1.53 & 4.08 & 4.95 & 3.64 &  2.30 \\ \hline
PLJP(CNN)                                   &  87.67 & \underline{55.21} & \underline{55.59} & \underline{54.37} & 86.05 & \underline{58.08} & 56.46 & 54.92 \\
PLJP(BERT)                                  & \textbf{94.18} & \textbf{74.65} & \textbf{76.23} & \textbf{74.84} & \underline{87.07} & \textbf{58.81} & \textbf{57.29} & \textbf{56.63} \\ \hline
\end{tabular}
\caption{Results of law article prediction, the best is \textbf{bolded} and the second best is \underline{underlined}.}
\label{tab:articleresults}
\end{table*}

\begin{table*}[t]
\small
    \centering
\begin{tabular}{lcccc|cccc}
\hline
\multicolumn{1}{c}{\multirow{2}{*}{Method}} & \multicolumn{4}{c|}{CJO22} & \multicolumn{4}{c}{CAIL2018} \\
\multicolumn{1}{c}{}                        & Acc   & Ma-P  & Ma-R  & Ma-F  & Acc  & Ma-P & Ma-R & Ma-F \\ \hline
CNN \citep{lecun1989backpropagation}        & 74.91 & 74.00 & 78.12 & 73.97 & 87.52 & 88.23 & 88.31 & 88.17 \\
BERT \citep{devlin2018bert}                 & 80.50 & 80.34 & 81.09 & 78.36 & 89.10 & 90.10 & 89.48 & 89.63 \\
Roberta \citep{liu2019roberta}              & 79.26 & 78.93 & 81.25 & 78.18 & 90.30 & 91.02 & 90.97 & 90.94 \\
TopJudge \citep{zhong2018legal}             & 76.67 & 74.00 & 77.40 & 74.62 & 87.31 & 88.68 & 87.84 & 88.20  \\
R-Former \citep{dong2021legal}              & 90.71 & \textbf{93.06} & \underline{88.66} & \textbf{89.82} &  \underline{91.54} &   \underline{91.61}  &  \textbf{91.96}  &  \textbf{91.58}         \\
LADAN \citep{xu2020distinguish}             & 79.64  & 48.43 & 44.13 & 46.18 & 88.09 & 90.12 & 88.82 & 89.47  \\
NeurJudge \citep{yue2021neurjudge}          & 71.85 &  69.37  & 71.09  & 68.66 & 82.13 & 82.71 & 82.30 & 82.36\\ 
EPM\citep{feng2022legal} & 83.49 & 80.36 & 83.29 & 81.87 & 91.20 & 90.81 & 89.99 & 90.46\\ 						
CTM\citep{liu2022augmenting} & 79.33 & 82.39 & 83.12 & 82.81 & 90.28 & 90.34 & 88.08 & 86.30 \\ 
Dav003          & 44.65  & 52.43 & 32.93  & 35.29  & 25.85 & 35.37 & 25.09 & 22.08 \\ 
3.5turbo         & 58.37  & 56.03  &  40.68 & 42.62 & 49.65 &  42.29
& 34.05 & 31.85\\ \hline
PLJP(CNN)                                   & \underline{91.62} & 83.43 & 84.88 & 83.40 & 91.49 & 81.80 & 83.95 & 80.06 \\
PLJP(BERT)                                  & \textbf{94.18} & \underline{90.25} & \textbf{88.67} & \underline{89.05} & \textbf{94.99} & \textbf{92.12} & \underline{91.10} & \underline{91.33}  \\ \hline
\end{tabular}
\caption{Results of charge prediction, the best is \textbf{bolded} and the second best is \underline{underlined}.}
\label{tab:chargeresults}
\end{table*}

\subsection{Datasets}

Following many influential LJP works \citep{zhong2018legal, xu2020distinguish, yue2021neurjudge, dong2021legal}, our experiment is conducted on the widely used and publicly available CAIL2018 dataset, which is a Chinese dataset in the context of People's Republic of China (PRC). This dataset consists of real-world cases, each of which includes a fact description accompanied by a complete judgment encompassing three labels: law articles, charges, and prison terms\footnote{Prison terms are divided into non-overlapping intervals.}.

To mitigate the potential data leakage during the training of LLMs, which were trained on corpora collected until September 2021, we have compiled a new dataset called CJO22. This dataset exclusively contains legal cases that occurred after 2022, sourced from the same origin as CAIL2018\footnote{\url{https://wenshu.court.gov.cn/}}. However, due to its limited size, the newly collected CJO22 dataset is inadequate for the training purposes of the domain models. Consequently, we utilize it solely as an additional test set. To facilitate meaningful comparisons, we retain only the labels that are common to both datasets, considering that the labels may not be entirely aligned.

Tab. \ref{tab:stat} shows the statistics of the processed datasets, and all the experiments are conducted on the same datasets. For CAIL2018 dataset, we randomly divide it into training set, validation set and test set according to the ratio of 8: 1: 1. 

The previous cases in the case database are sampled from the training dataset, and we set the amount to 4000.

\begin{table*}[t]
\small
    \centering
\begin{tabular}{lcccc|cccc}
\hline
\multicolumn{1}{c}{\multirow{2}{*}{Method}} & \multicolumn{4}{c|}{CJO22} & \multicolumn{4}{c}{CAIL2018} \\
\multicolumn{1}{c}{}                        & Acc   & Ma-P  & Ma-R  & Ma-F  & Acc  & Ma-P & Ma-R & Ma-F \\ \hline
CNN \citep{lecun1989backpropagation}        & 27.38 & 18.48 & 17.51 & 17.44 & 34.42 & 32.22 & 30.53 & 31.05  \\
BERT \citep{devlin2018bert}                 & 36.80 & 29.83 & 27.50 & 27.03 & 40.00 & 37.53 & 33.66 & 33.58 \\
Roberta \citep{liu2019roberta}              & 29.74 & 24.73 & 24.76 & 23.22 & \underline{40.84} & \underline{38.62} & \textbf{38.55} & \textbf{38.50} \\
TopJudge \citep{zhong2018legal}             & 27.14 & 19.76 & 17.69 & 17.94 & 35.54 & 33.55 & 31.08 & 32.00 \\
R-Former \citep{dong2021legal}              & \underline{38.63}  & \underline{32.63} & \underline{32.76} & \underline{29.51} &  40.70  &  36.09 & 36.76 & 35.04  \\
LADAN \citep{xu2020distinguish}             & 33.69 &  26.40  & 22.94 & 24.55 &  38.03 & 33.66 & 30.08 & 31.77 \\
NeurJudge \citep{yue2021neurjudge}          & 26.80 &  26.81 & 26.85 & 25.97  &  33.53  & 36.46 & 37.26 & 36.53    \\ 
EPM\citep{feng2022legal} & 36.91 & 30.65 & 31.61 & 30.20 & 40.25 & 37.96	& 37.00	& 37.34\\ 						
CTM\citep{liu2022augmenting} & 36.81	& 27.10 & 25.96 & 26.46 & 39.56 & 38.66 & \underline{38.02} & \underline{37.84} \\ 
Dav003          & 0.47 &  5.56 & 0.21  & 0.41  & 0.68 & 10.38 & 0.49 & 0.94 \\ 
3.5turbo         & 1.40  & 1.16  &  1.07  &  1.11  & 1.02 & 2.71 & 1.13 & 1.15 \\ \hline
PLJP(CNN)                                   & 36.51 & 20.21 & 21.44 & 20.07 & 40.81 & 32.77 & 35.59 & 25.71 \\
PLJP(BERT)                                  & \textbf{43.52} & \textbf{33.37} & \textbf{35.67} & \textbf{31.98} & \textbf{48.72} & \textbf{42.64} & 36.80 & 35.43 \\ \hline
\end{tabular}
\caption{Results of prison term prediction, the best is \textbf{bolded} and the second best is \underline{underlined}.}
\label{tab:prisonresults}
\end{table*}

\begin{table*}[ht]
\centering
\small
\begin{tabular}{lcccccc|cccccc}
\hline
\multicolumn{1}{c}{\multirow{3}{*}{Method}} & \multicolumn{6}{c|}{CJO22}                                                                      & \multicolumn{6}{c}{CAIL2018}                                                                   \\
\multicolumn{1}{c}{}                        & \multicolumn{2}{c}{Law Article} & \multicolumn{2}{c}{Charge} & \multicolumn{2}{c|}{Prison Term} & \multicolumn{2}{c}{Law Article} & \multicolumn{2}{c}{Charge} & \multicolumn{2}{c}{Prison Term} \\
\multicolumn{1}{c}{}                        & Acc            & Ma-F           & Acc         & Ma-F         & Acc            & Ma-F            & Acc            & Ma-F           & Acc         & Ma-F         & Acc            & Ma-F           \\ \hline
w/o p                                       &   54.65    &   28.32   &  83.48    &   \underline{76.33}     &  35.81    &   20.84        & 85.03  &   51.54   & 85.03 &  70.07  &   32.31   &   22.58    \\
w/o c               &  45.34      &   40.22   &   42.32    &  41.85  &  32.55  &   20.26    & 67.35   &  46.65    & 72.79   &  60.34    &   26.53   &   13.66   \\
w/o d                                       &  \textbf{94.18}      &  \textbf{74.84}    & 85.58   &  70.50   & \underline{39.53}      &  20.31        &  \underline{87.07}   &   56.63    &   \underline{87.41}   &    73.45   &    38.09   &  21.44    \\
w/o r                                       &   88.13      &   58.75   &  \underline{87.67}  &   74.83    &  36.27     &  \underline{23.70}        &   86.05  &   \underline{58.26}   &  86.73  &  \underline{77.53}   &  38.10    &  21.70        \\ 
 w/ e & \underline{90.70} & \underline{67.90} & 80.70 & 66.53& 35.35 & 20.21& \textbf{89.80} & \textbf{61.64}& 85.37 & 68.48 & \underline{38.44} & \underline{23.14} \\
\hline
PLJP                                        &    \textbf{94.18}   &   \textbf{74.84}   &   \textbf{94.18}   &  \textbf{89.05}   &  \textbf{43.52}     &   \textbf{31.98}       &  \underline{87.07}  &    56.63   &  \textbf{94.99}  &  \textbf{91.33}   &   \textbf{48.72}   &   \textbf{35.43}     \\ \hline
\end{tabular}
\caption{Results of ablation experiments, the best is \textbf{bolded} and the second best is \underline{underlined}.}
\label{tab:ablation}
\end{table*}

\begin{figure}[t]
\centering
\includegraphics[width=0.48\textwidth]{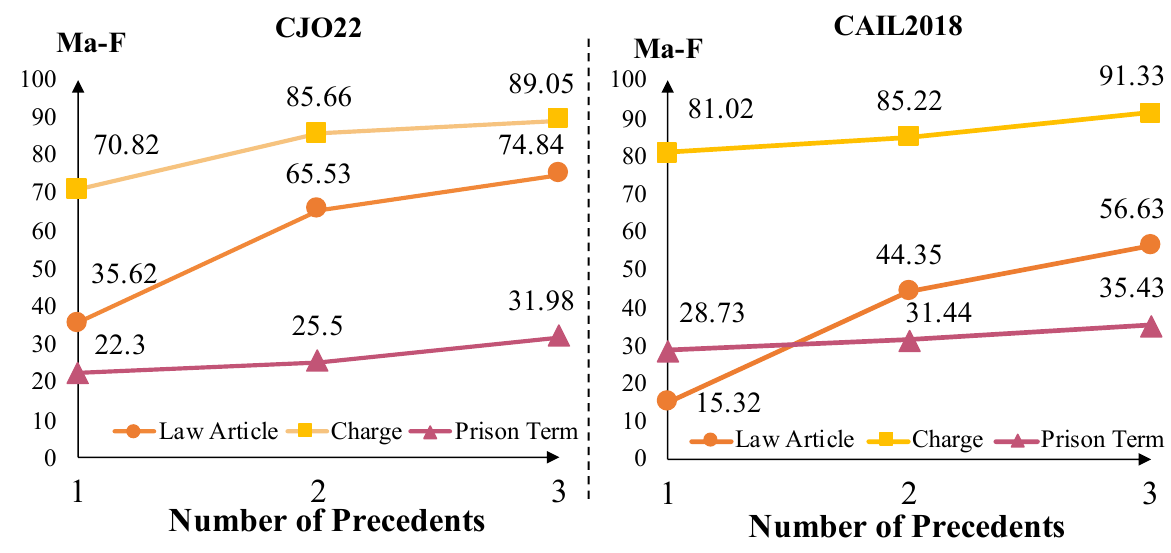}
\caption{The Ma-F of PLJP with different number of precedents.}
\label{fig:precedent}
\vspace{-10pt}
\end{figure}

\subsection{Baselines}
For domain-specific LJP baselines, we implement the following for comparison:

\textbf{CNN} \citep{lecun1989backpropagation} extracts text features through convolutional operations with different kernels for text classification; \textbf{BERT}\citep{devlin2018bert} is a pre-trained language model and can be easily fine-tuned on the downstream tasks; \textbf{TopJudge} \citep{zhong2018legal} use multi-task learning and capture the dependencies among the three sub-task in LJP; \textbf{NeurJudge} \citep{yue2021neurjudge} splits the fact description into different parts for making predictions; \textbf{R-Former} \citep{dong2021legal} formalizes LJP as a node classification problem over a global consistency graph and relational learning is introduced; \textbf{LADAN} \citep{xu2020distinguish} uses graph distillation to extract discriminative features of the fact
\textbf{Retri-BERT} \citep{chalkidis2023retrieval} retrieves similar documents to augment the input document representation for multi-label text classification; \textbf{EPM} \citep{feng2022legal} locates event-related information essential for judgment while utilizing cross-task consistency constraints among the subtasks; \textbf{CTM} \citep{liu2022augmenting} establishes a LJP framework with case triple modeling from contrastive case relations.

We use the LLM baselines as follows\footnote{We give a fixed example in the prompt to help the LLMs understand the tasks.}:
\textbf{Dav003} means the text-davinci-003, \textbf{3.5turbo} means the gpt-3.5-turbo. These LLMs are both from the GPT-3.5 family, released by OpenAI and can understand and generate complex natural language\footnote{\url{https://platform.openai.com/docs/models}}.

\begin{figure*}[ht]
    \centering
    \includegraphics[width=1\textwidth]{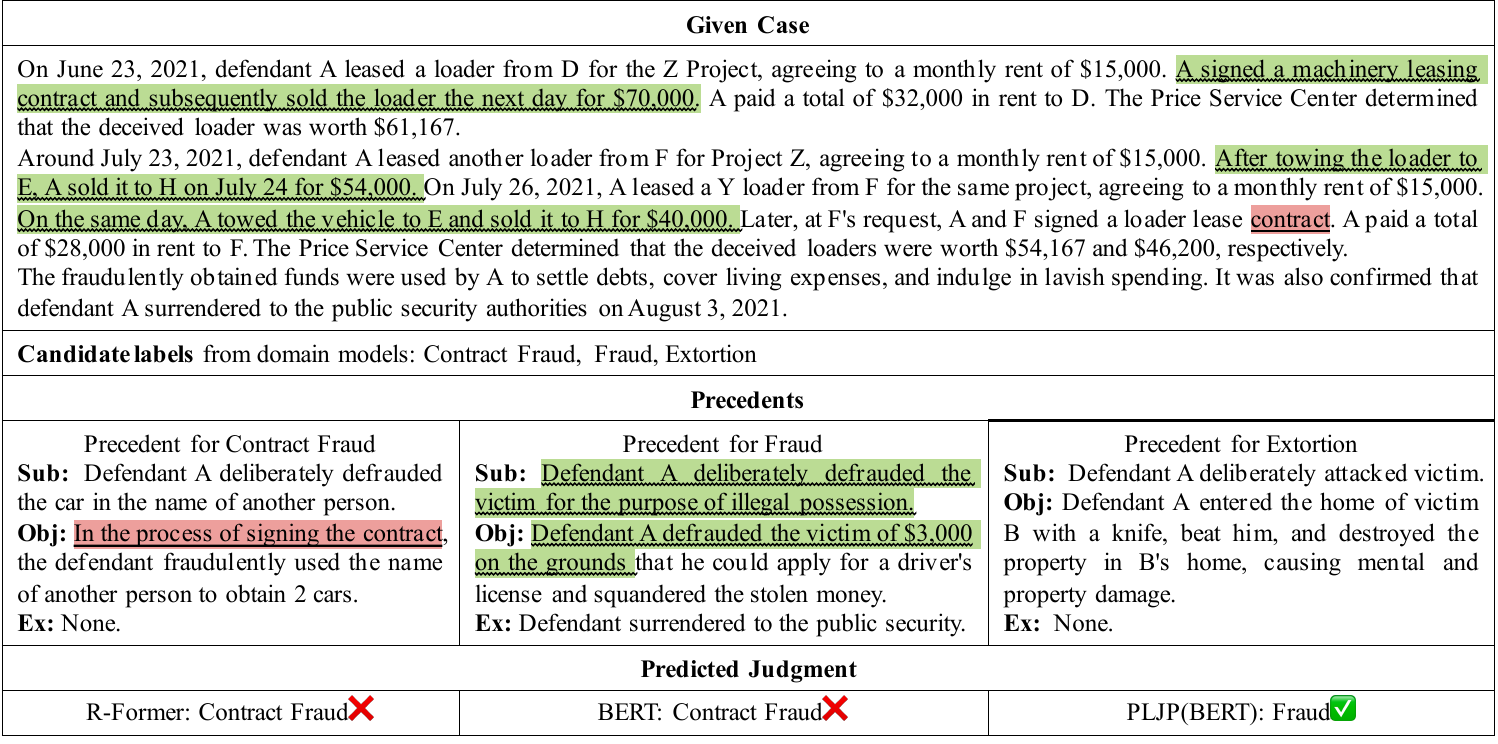}
    \caption{The charge prediction of a given case. The \uwave{green} parts are useful information for prediction, while the \underline{red} parts are content that can be confused by the domain models.}
    \vspace{-10pt}
    \label{fig:showcase}
\end{figure*}

For PLJP, we take the CNN and BERT as the predictive models, and take the text-davinci-003 as the implementation of the LLM, named as PLJP(CNN) and PLJP(BERT). The top-k accuracy of CNN and BERT is shown in the Appendix. Considering the length limit of the prompt, we set the number of precedents to 3.

We also do ablation experiments as follows:
\textbf{PLJP w/o p} refers to the removal of \underline{p}recedents, and the prediction of labels is done solely based on the candidate labels using the LLM;
\textbf{PLJP w/o c} denotes we remove the \underline{c}andidate labels and predict the label only with the fact description and precedents; \textbf{PLJP w/o d} means we predict the three labels independently instead of considering the \underline{d}ependencies among the three subtasks; \textbf{PLJP w/o r} denotes we find precedents based the raw fact instead of from the \underline{r}eorganized fact; \textbf{PLJP w/ e} means we let the LLMs generate the \underline{e}xplanation of the prediction as well.

In the ablation study, PLJP means PLJP(BERT).

\subsection{Experiment Settings}
Here we describe the implementation of PLJP in our experiments. Note all the LLMs and domain models are replaceable in the PLJP framework. 

In the experiments, for the LLMs, we directly use the APIs provided by OpenAI.
For the domain models, we use the unsupervised dense retrieval model \citep{izacard2022unsupervised} in precedent retrieval, which gets the precedents from the case database according to the reorganized facts. For other domain models such as TopJudge and NeurJudge, we use the training settings from the original paper.

For the metrics, we employ Accuracy (Acc), Macro-Precision (Ma-P), Macro-Recall (Ma-R) and Macro-F1 (Ma-F).

\subsection{Experiment Results}
We analyze the experimental results in this section.
\paragraph{Result of judgment prediction:} From Tab. \ref{tab:articleresults}, Tab. \ref{tab:chargeresults} and Tab. \ref{tab:prisonresults}, we have the following observations:
1) The LLMs perform not well in the prediction tasks alone, especially when the label has no actual meaning (e.g., the index of the law article and prison term).
2) By applying our PLJP framework with the collaboration of LLMs and domain models, the simple models (e.g., CNN, BERT) gain significant improvement.
3) The model performance on CJO22 is lower than that on CAIL2018, which shows the challenge of the newly constructed test set.
4) PLJP(BERT) achieves the best performance in almost all the metric evaluation metrics in both CAIL2018 and CJO22 test sets, which proves the effectiveness of the PLJP.
5) Compared to the prediction of the law article and charge, the prediction of prison term is still a more challenging task.
6) The reported results of the LJP baselines are not as good as the original papers, this may be because we keep all the low-frequency labels instead of removing them as the original papers did.

\paragraph{Results of ablation experiment:} From Tab. \ref{tab:ablation}, we can conclude that:
1) The performance gap of the PLJP w/o p and PLJP demonstrates the effects of the precedents.
2) The results of PLJP w/o c prove the importance of the candidate labels.
3) Considering the topological dependence of the three sub-tasks benefits the model performance as PLJP w/o d shows.
4) When we use the raw fact instead of the reorganized fact, the performance drops (e.g., the Acc of prison term in CJO22 drops from 45.32\% to 36.27\%).
5) If we force the LLMs to generate the explanation of the prediction, the performance also drops a bit. We put cases with explanations in the Appendix.

From Fig. \ref{fig:precedent}, we can find that the performance of PLJP improves as the number of precedents increases, which also proves the effectiveness of injecting precedents into the LJP.

\subsection{Case Study}

Fig. \ref{fig:showcase} shows an intuitive comparison among the three methods in the process of charge prediction. Based on the fact description of the given case, the domain models provide candidate charges with the corresponding precedents. As the case shows, the defendant made fraud by selling the cars that were rented from other people. However, since there contains ``contract'' in the fact description, baselines (e.g., R-Former and BERT) can be misled and predict the wrong charge of ``Contract Fraud''. Through an in-context precedent comprehension by the LLMs, PLJP(BERT) distinguishes the differences among the precedents and the given case (e.g., the crime does not occur during the contracting process, and the contract is only a means to commit the crime), and give the right result of ``Fraud''.

\section{Conclusion and Future Work}
In this paper, we address the important task of legal judgment prediction (LJP) by taking precedents into consideration. We propose a novel framework called precedent-enhanced legal judgment prediction (PLJP), which combines the strength of both LLMs and domain models to better utilize (e.g., retrieve and comprehend) the precedents. Experiments on the real-world dataset prove the effectiveness of the PLJP.

Based on the PLJP, in the future, we can explore the following directions: 
1) Develop methods to identify and mitigate any biases that could affect the predictions and ensure fair and equitable outcomes.
2) Validate the effectiveness of LLM and domain collaboration in other vertical domains such as medicine and education.

\subsection{Ethical Discussion}
With the increasing adoption of Legal AI in the field of legal justice, there has been a growing awareness of the ethical implications involved. The potential for even minor errors or biases in AI-powered systems can lead to significant consequences.

In light of these concerns, we have to claim that our work is an algorithmic exploration and will not be directly used in court so far. Our goal is to provide suggestions to judges rather than making final judgments without human intervention. In practical use, human judges should be the final safeguard to protect justice fairness. In the future, we plan to study how to identify and mitigate potential biases to ensure the fairness of the model.

\section{Limitations}
In this section, we discuss the limitations of our works as follow:

$\bullet$ We only interact with the LLMs one round per time. The LLMs are capable of multi-round interaction (e.g., Though of Chains), which may help the LLM to better understand the LJP task.

$\bullet$ We validate the effectiveness of LLM and domain model collaboration in the legal domain. It's worthwhile to explore such collaboration in other vertical domains such as medicine and education, as well as in other legal datasets (e.g., the datasets from the Common Law system).

\normalem
\bibliography{anthology,custom}
\bibliographystyle{acl_natbib}

\appendix

\section{Appendices}

\subsection{Top-k Accuracy}

\begin{figure}[H]
\centering
\includegraphics[width=0.4\textwidth]{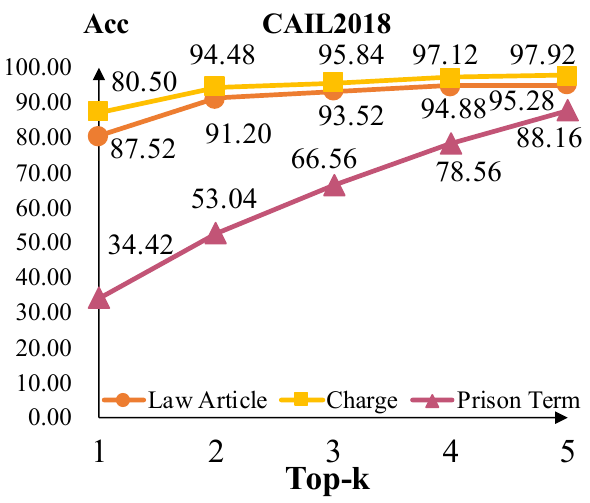}
\caption{The top-k accuracy of CNN on CAIL dataset.}
\end{figure}

\begin{figure}[H]
\centering
\includegraphics[width=0.4\textwidth]{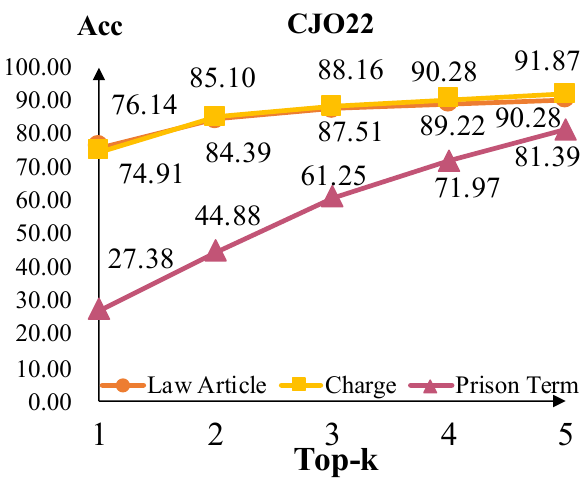}
\caption{The top-k accuracy of CNN on CJO22 dataset.}
\end{figure}

\begin{figure}[H]
\centering
\includegraphics[width=0.4\textwidth]{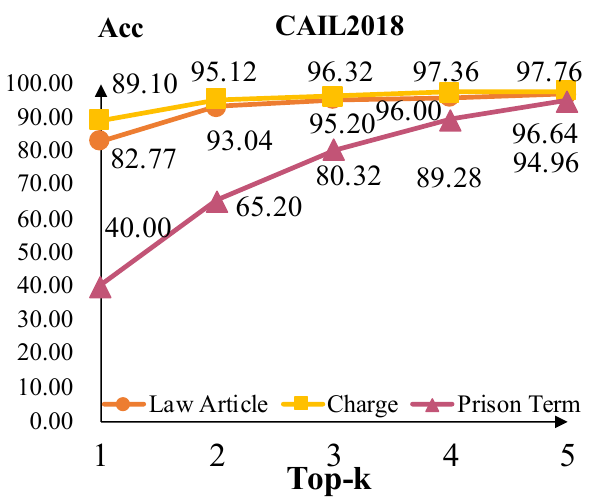}
\caption{The top-k accuracy of BERT on CAIL dataset.}
\end{figure}

\begin{figure}[H]
\centering
\includegraphics[width=0.4\textwidth]{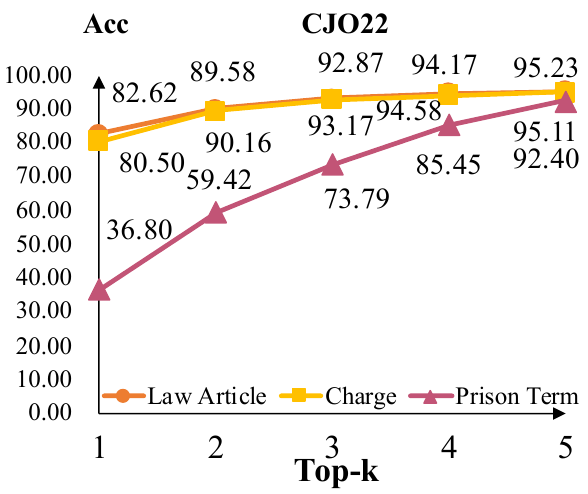}
\caption{The top-k accuracy of BERT on CJO22 dataset.}
\end{figure}

\subsection{More Show Cases}


\begin{figure*}[ht]
    \centering
    \includegraphics[width=1\textwidth]{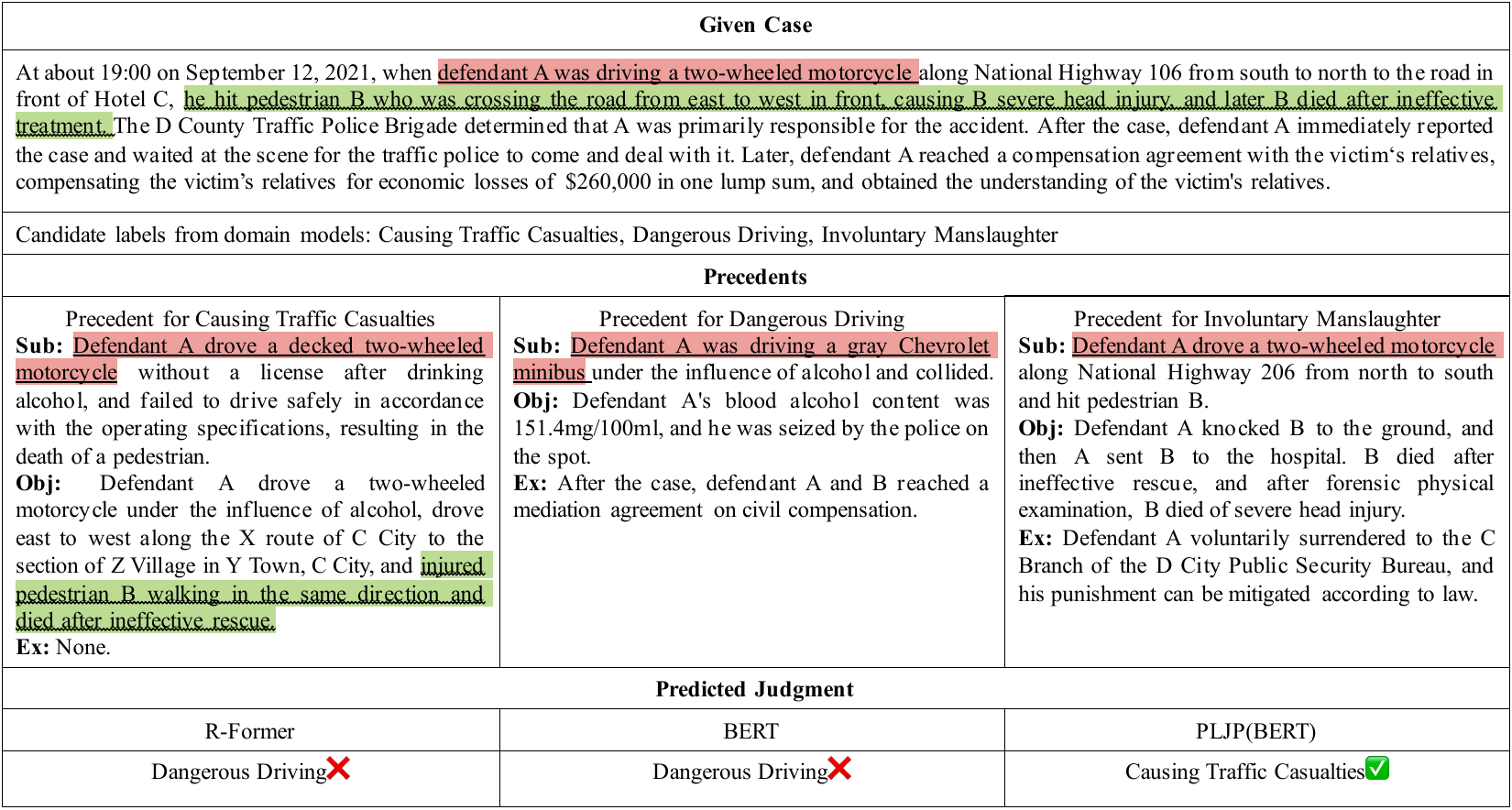}
    \caption{More case 1.}
    \label{fig:morecase1}
\end{figure*}

\begin{figure*}[t]
    \centering
    \includegraphics[width=1\textwidth]{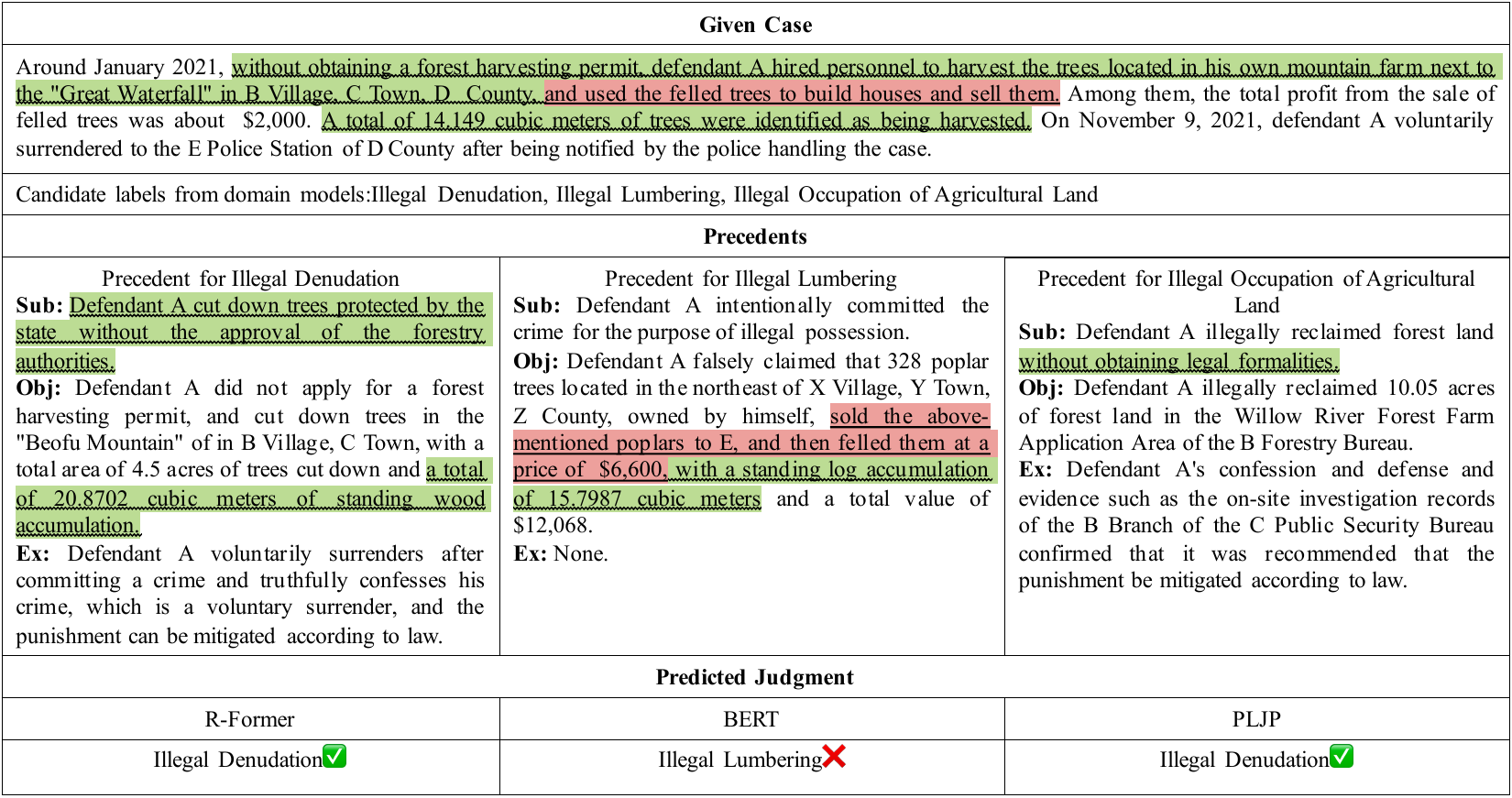}
    \caption{More case 2.}
    \label{fig:morecase2}
\end{figure*}

\begin{figure*}[t]
    \centering
    \includegraphics[width=1\textwidth]{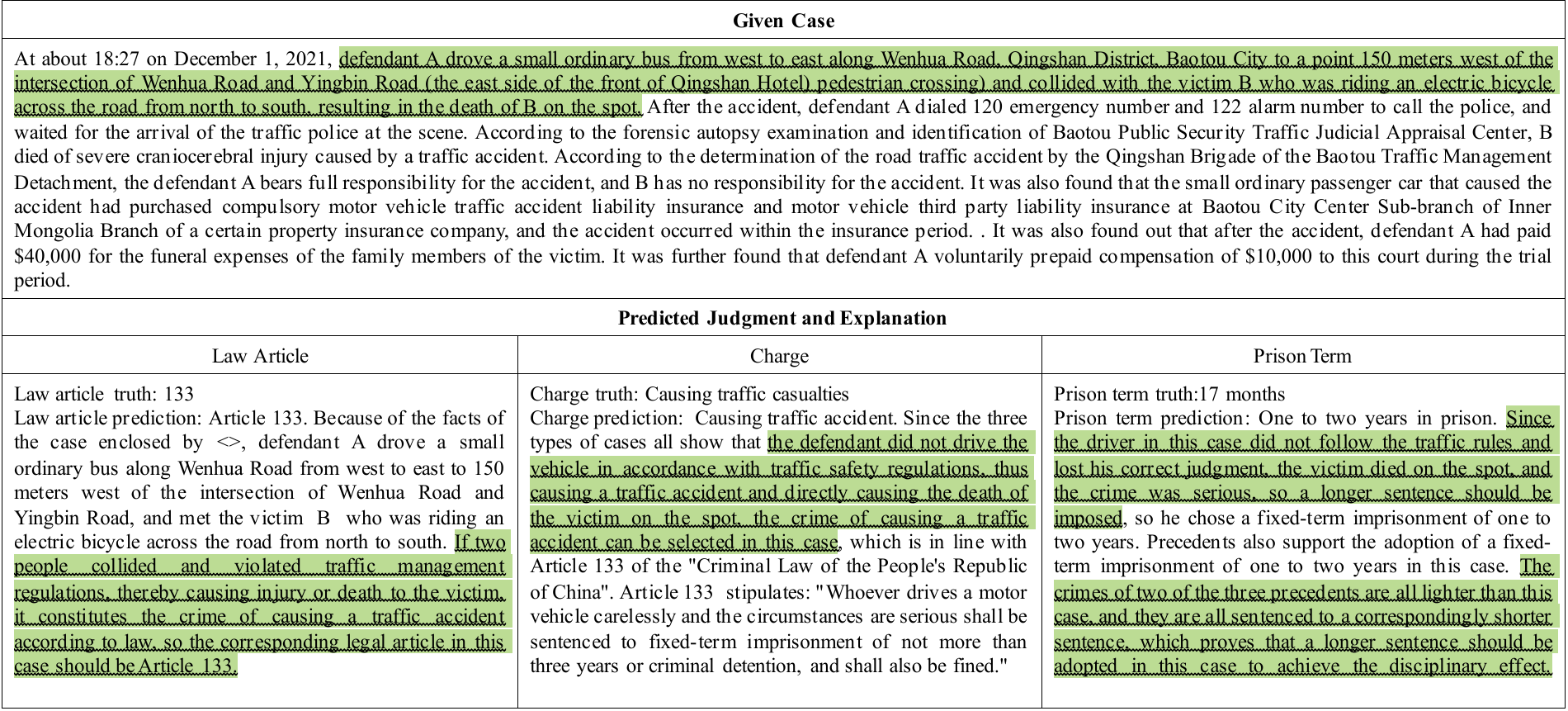}
    \caption{Case with explanations 1.}
    \label{fig:casewq1}
\end{figure*}

\begin{figure*}[t]
    \centering
    \includegraphics[width=1\textwidth]{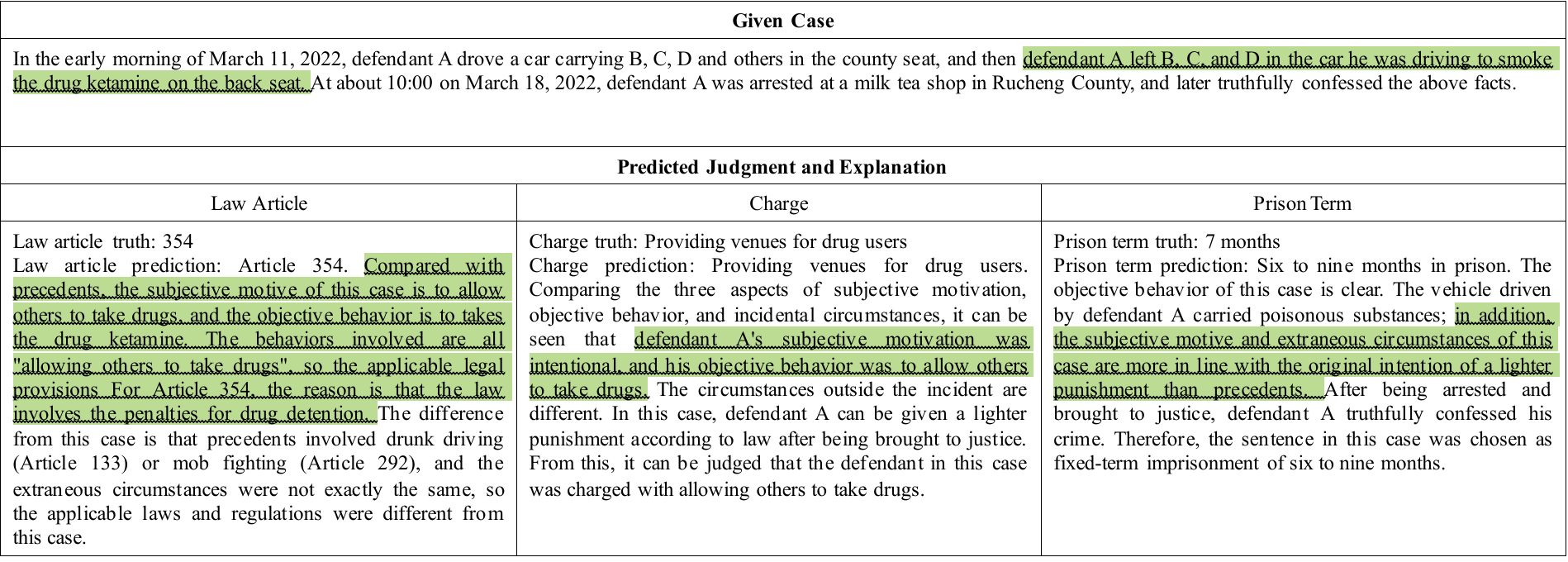}
    \caption{Case with explanations 2.}
    \label{fig:casewq2}
\end{figure*}

\begin{figure*}[t]
    \centering
    \includegraphics[width=1\textwidth]{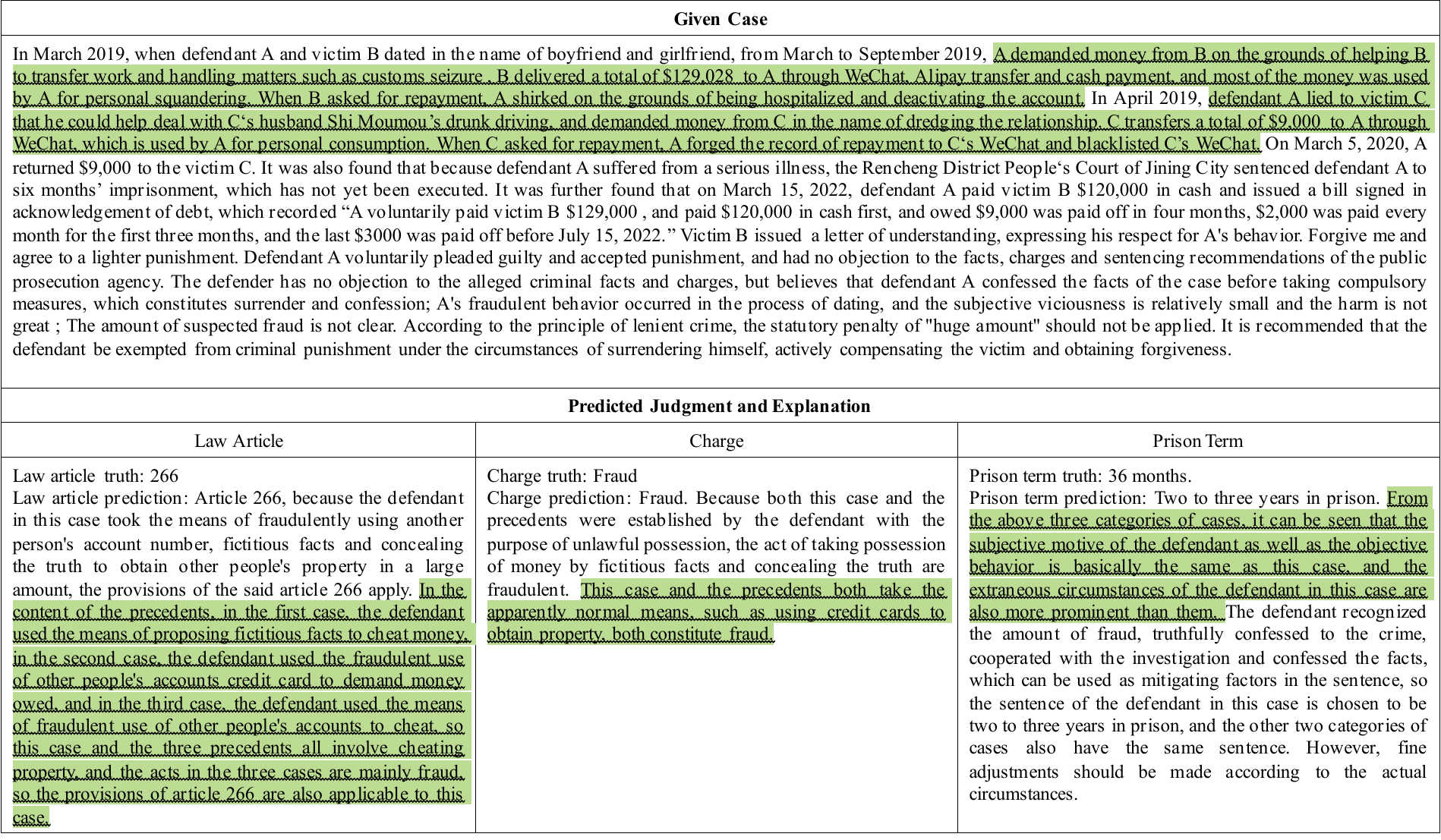}
    \caption{Case with explanations 3.}
    \label{fig:casewq3}
\end{figure*}

\end{document}